\documentclass[11pt,letterpaper]{article}
\usepackage[hyperref]{acl2020}
\usepackage{times}
\usepackage{latexsym}

\usepackage{microtype}

\usepackage{graphicx}
\usepackage{breakurl}

\usepackage{multirow}

\usepackage[totalwidth=158mm, totalheight=248mm]{geometry}

\aclfinalcopy 
\def\aclpaperid{1014} 

\newcommand{\minisection}[1]{\noindent{\bf {#1}}}
\newcommand{\argmin}{\mathop{\rm arg~min}\limits}

\title{Revisiting the Context Window for Cross-lingual Word Embeddings}

\author{Ryokan Ri and Yoshimasa Tsuruoka \\
  The University of Tokyo\\
  7-3-1 Hongo, Bunkyo-ku, Tokyo, Japan \\
  {\tt \{li0123,tsuruoka\}@logos.t.u-tokyo.ac.jp} \\}

\date{}

\begin{document}

\maketitle
\begin{abstract}
Existing approaches to mapping-based cross-lingual word embeddings are based on the assumption that the source and target embedding spaces are structurally similar.
The structures of embedding spaces largely depend on the co-occurrence statistics of each word, which the choice of context window determines.
Despite this obvious connection between the context window and mapping-based cross-lingual embeddings, their relationship has been underexplored in prior work.
In this work, we provide a thorough evaluation, in various languages, domains, and tasks, of bilingual embeddings trained with different context windows.
The highlight of our findings is that increasing the size of both the source and target window sizes improves the performance of bilingual lexicon induction, especially the performance on frequent nouns.

\end{abstract}

\section{Introduction}

Cross-lingual word embeddings can capture word semantics invariant among multiple languages, and facilitate cross-lingual transfer for low-resource languages \citep{Ruder:2019up}.
Recent research has focused on {\it mapping-based} methods, which find a linear transformation from the source to target embedding spaces \citep{Mikolov:2013tp, Artetxe:2016wb, Lample:2018wg}.
Learning a linear transformation is based on a strong assumption that the two embedding spaces are structurally similar or isometric.

The structure of word embeddings heavily depends on the co-occurrence information of words \citep{Turney:2010tn, Baroni:2014vd}, {\it i.e.}, word embeddings are computed by counting other words that appear in a specific context window of each word.
The choice of context window changes the co-occurrence statistics of words and thus is crucial to determine the structure of an embedding space.
For example, it has been known that an embedding space trained with a smaller linear window captures functional similarities, while a larger window captures topical similarities \citep{Levy:2014we}.
Despite this important relationship between the choice of context window and the structure of embedding space, how the choice of context window affects the structural similarity of two embedding spaces has not been fully explored yet.

In this paper, we attempt to deepen the understanding of cross-lingual word embeddings from the perspective of the choice of the context window through carefully designed experiments.
We experiment with a variety of settings, with different domains and languages. We train monolingual word embeddings varying the context window sizes, align them with a mapping-based method, and then evaluate them with both intrinsic and downstream cross-lingual transfer tasks.
Our research questions and the summary of the findings are as follows:

\minisection{RQ1: What kind of context windows produces a better alignment of two embedding spaces?}
Our result shows that increasing the window sizes of both the source and target embeddings improves the accuracy of bilingual dictionary induction consistently regardless of the domains of the source and target corpora.
Our fine-grained analysis reveals that frequent nouns receive the most benefit from larger context sizes.

\minisection{RQ2. In downstream cross-lingual transfer, do the context windows that perform well on the source language also perform well on the target languages?}
No.
We find that even when some context window performs well on the source language task, that is often not the best choice for the target language.
The general tendency is that broader context windows produce better performance for the target languages.



\section{Background and Related Work}
\subsection{Context Window of Word Embeddings}
Word embeddings are computed from the co-occurrence information of words, {\it i.e.}, context words that appear around a given word. The embedding algorithm used in this work is the skip-gram with negative sampling \citep{NIPS2013_5021}. In the skip-gram model, each word $w$ in the vocabulary $W$ is associated with a word vector $v_{w}$ and a context vector $c_{w}$.\footnote{Conceptually, the word and context vocabularies are regarded as separated, but for simplicity, we assume that they share the vocabulary.} The objective is to maximize the dot-product $v_{w_t} \cdot c_{w_c}$ for the observed word-context pairs $(w_{t}, w_{c})$, and to minimize the dot-product for negative examples.

The most common type of context is a linear window. When the window size is set to $k$, the context words of a target word $w_{t}$ in a sentence $[w_1, w_2, ..., w_{t}, ...w_{L}]$ are $[w_{t-k}, ..., w_{t-1}, w_{t+1}, ..., w_{t+k}]$.
The choice of context is crucial to the resulting embeddings as it will change the co-occurrence statistics associated with each target word.
Table \ref{tb:nn_example} demonstrates the effect of the context window size on the nearest neighbor structure of embedding space; with a small window size, the resulting embeddings capture functional similarity, while with a larger window size, the embeddings capture topical similarities.

Among the other types of context windows that have been explored by researchers are linear windows enriched with positional information \citep{Levy:2014wb, Ling:2015us, Li:2017ta}, syntactically informed context windows based on dependency trees \citep{Levy:2014we, Li:2017ta}, and one that dynamically weights the surrounding words with the attention mechanism \citep{Ling:2015vv}.
In this paper, we mainly discuss the most common linear window and investigate how the choice of the window size affects the isomorphism of two embedding spaces and the performance of cross-lingual transfer.

\begin{table}[]
  \begin{tabular}{|l|c|c|} \hline
    Query word & window size 1 & window size 10 \\ \hline \hline
    %

               & phrases       & word           \\ \cline{2-3}
               & loanwords     & phrases        \\ \cline{2-3}
    words      & morphemes     & phrase         \\ \cline{2-3}
               & verses        & ungrammatical  \\ \cline{2-3}
               & phonemes      & homographs     \\ \hline

                & synchronic      & totemism        \\ \cline{2-3}
                & mechanistic     & typology        \\ \cline{2-3}
    typological & numerological   & categorizations \\ \cline{2-3}
                & architectonic   & dialectology    \\ \cline{2-3}
                & dialectical     & fusional        \\ \hline
  \end{tabular}
  \caption{The top-5 nearest neighbors in English embedding spaces trained with different context windows in our experiment. The smaller window size captures functional similarities ({\it -s, -cal, -ic}), while the larger captures topical similarities.}
  \label{tb:nn_example}
  \vspace{-5mm}
\end{table}



\subsection{Cross-lingual Word Embeddings}
Cross-lingual word embeddings aim to learn a shared semantic space in multiple languages.
One promising solution is to jointly train the source and target embedding, so-called {\it joint methods}, by exploiting cross-lingual supervision signals in the form of word dictionaries \citep{Duong:2016vw}, parallel corpora \citep{Gouws:2015ws, Luong:2015uh}, document-aligned corpora \citep{Vulic:2016tj}.

Another line of research is off-line mapping-based approaches \citep{Ruder:2019up}, where monolingual embeddings are independently trained in multiple languages, and a post-hoc alignment matrix is learned to align the embedding spaces with a seed word dictionary \citep{Mikolov:2013tp, Xing:2015ur, Artetxe:2016wb}, with only a little supervision such as identical strings or numerals \citep{Artetxe:2017ht, Smith:2017vw}, or even in a completely unsupervised manner \citep{Lample:2018wg, Artetxe:2018vx}. Mapping-based approaches have recently been popularized by their cheaper computational cost compared to joint approaches, as they can make use of pre-trained monolingual word embeddings.

The assumption behind the mapping-based methods is the isomorphism of monolingual embedding spaces, {\it i.e.}, the embedding spaces are structurally similar, or the nearest neighbor graphs from the different languages are approximately isomorphic \citep{Sogaard:2018vo}.
Considering that the structures of the monolingual embedding spaces are closely related to the choice of the context window, it is natural to expect that the context window has a considerable impact on the performance of mapping-based bilingual word embeddings.

However, most existing work has not provided empirical results on the effect of the context window on cross-lingual embeddings, as their focus is on how to learn a mapping between the two embedding spaces.
In order to shed light on the effect of the context window on cross-lingual embeddings, we trained cross-lingual embeddings with different context windows, and carefully analyzed the implications of their varying performance on both intrinsic and extrinsic tasks.

\section{Experimental Design}
\subsection{Training Monolingual Embeddings}
The experiment is designed to deal with multiple settings to fully understand the effect of the context window.

\minisection{Languages.} As the target language, we choose English ({\textsc en}) because of its richness of resources, and as the source languages, we choose French ({\textsc fr}), German ({\textsc de}), Russian ({\textsc ru}), Japanese ({\textsc ja}), taking into account the typological variety and availability of evaluation resource.

Note that the language pairs analyzed in this paper are limited to those including English, and there is a possibility that some results may not generalize to other language pairs.

\minisection{Corpus for Training Word Embeddings.} To train the monolingual embeddings, we use the Wikipedia Comparable Corpora\footnote{\url{https://linguatools.org/tools/corpora/wikipedia-comparable-corpora/}}.
We choose comparable corpora for the main analysis in order to accentuate the effect of context window by setting an ideal situation for training cross-lingual embeddings.

We also experiment with different domain settings, where we use corpora from the news domain\footnote{https://wortschatz.uni-leipzig.de/en/download} for the source languages, because the isomorphism assumption is shown to be very sensitive to the domains of the source and target corpora \citep{Sogaard:2018vo}. We refer to those results when we are interested in whether the same trend with respect to context window can be observed in the different domain settings.

For the size of the data, to simulate the setting of transferring from a low-resource language to a high-resource language, we use 5M sentences for the target language (English), and 1M sentences for the source languages.\footnote{We also experimented with very low-resource settings, where the source corpus size is set to 100K, but the results showed similar trends to the 1M setting, and thus we only include the result of the 1M settings in this paper.}

\minisection{Context Window.} Since we want to measure the effect of the context window size, we vary the window size among 1, 2, 3, 4, 5, 7, 10, 15, and 20.

Besides the linear window, we also experimented with the unbound dependency context \citep{Li:2017ta}, where we extract context words that are the head, modifiers, and siblings in a dependency tree.
Our initial motivation was that, while the linear context is directly affected by different word orders, the dependency context can mitigate the effect of language differences, and thus may produce better cross-lingual embeddings.
However, the performance of the dependency context turned out to be always in the middle between smaller and larger linear windows, and we found nothing notable. Therefore, the following analysis only focuses on the results of the linear context window.

\minisection{Implementation of Word2Vec.}
Note that some common existing implementations of the skip-gram may obfuscate the effect of the window size.
The original C implementation of {\tt word2vec} and its python implementation {\tt Gensim}\footnote{\url{https://radimrehurek.com/gensim/}} adopt a dynamic window mechanism where the window size is uniformly sampled between 1 and the specified window size for each target word \citep{Mikolov:2013wc}.
Also, those implementations remove frequent tokens by subsampling {\it before} extracting word-context pairs (so-called ``dirty'' subsampling) \citep{Levy:2015us}, which enlarges the context size in effect.
Our experiment is based on {\tt word2vecf},\footnote{\url{https://bitbucket.org/yoavgo/word2vecf/src/default/}} which takes arbitrary word-context pairs as input. We extract word-context pairs from a fixed window size and afterward perform subsampling.

\begin{figure*}[t]
\vspace{-3mm}
\begin{center}
  \includegraphics[width=15cm]{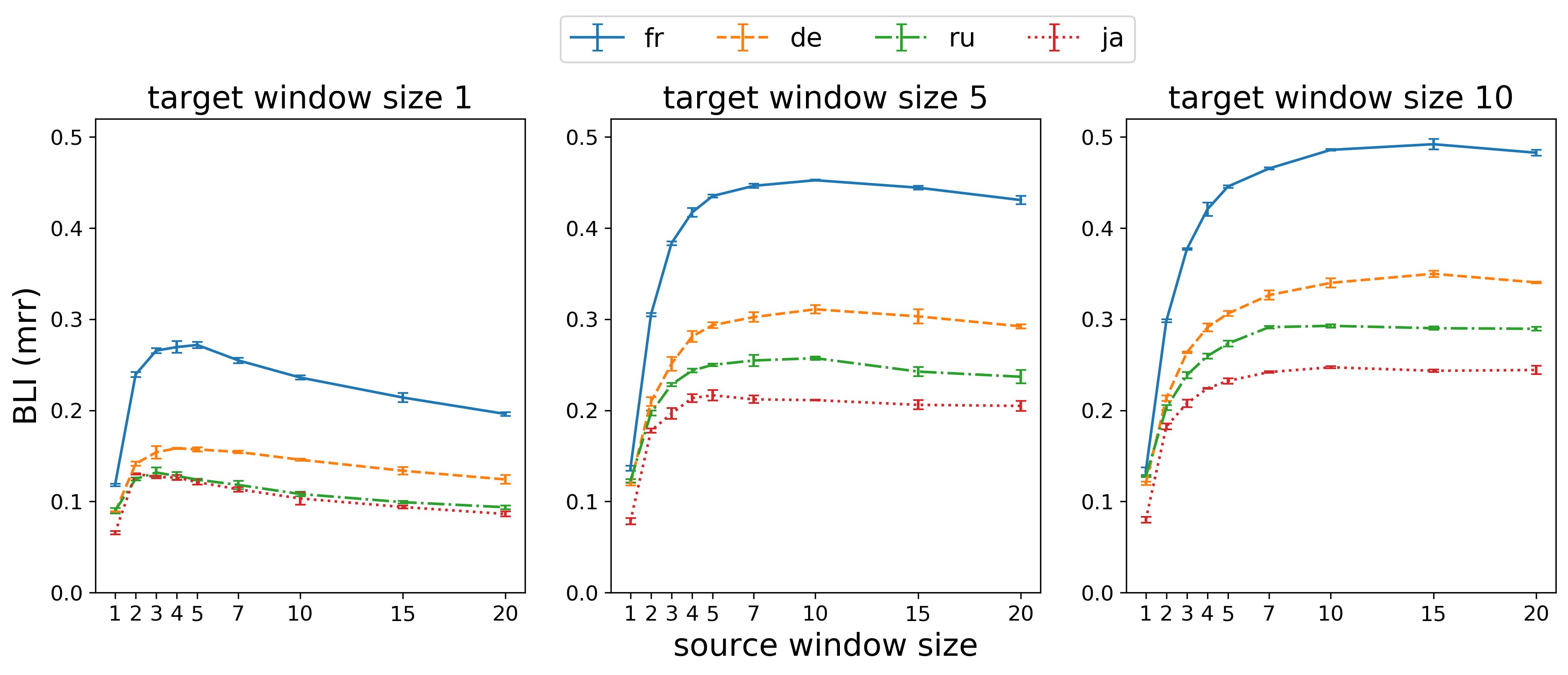}
  \caption{BLI performance in the comparable setting. The target window size is fixed and the source window size is varied.}
  \label{wiki_comp_fixed_window}
\end{center}
\vspace{-5mm}
\end{figure*}

We train 300-dimensional embeddings. For details on the hyperparameters, we refer the readers to Appendix \ref{appendix:WE}.

\subsection{Aligning Monolingual Embeddings}
After training monolingual embeddings in the source and target languages, we align them with a mapping-based algorithm.
To induce a alignment matrix $W$ for the source and target embeddings $x, y$,
we use a simple supervised method of solving the Procrustes problem $\argmin_{W} \sum_{i=1}^{m} \left\|W x_{i}-y_{i}\right\|^{2}$, with a training word dictionary ${(x_i, y_i)^{m}_{i=1}}$ \citep{Mikolov:2013tp},
with the orthogonality constraint on $W$, length normalization and mean-centering as preprocessing for the source and target embeddings \citep{Artetxe:2016wb}.

The word dictionaries are automatically created by using Google Translate. \footnote{https://translate.google.com/ (October 2019)} We translate all words in our English vocabulary into the source languages and filter out words that do not exist in the source vocabularies. We also perform this process in the opposite direction (translated from the source languages into English), and take the union of the two corresponding dictionaries. We then randomly select 5K tuples for training and 2K for testing.
Although using word dictionaries automatically derived from a system is currently a common practice in this field, it should be acknowledged that this may sometimes pose problems: the generated dictionaries are noisy, and the definition of word translation is unclear ({\it e.g., } how do we handle polysemy?). It can hinder valid comparisons between systems or detailed analysis of them, and should be addressed in future research.

For each setting, we train three pairs of aligned embeddings with different random seeds in the monolingual embedding training, as training word embeddings is known to be unstable and different runs result in different nearest neighbors \citep{wendlandt-etal-2018-factors}. The following results are presented with their averages and standard deviations.

\section{Bilingual Lexicon Induction}
We first evaluate the learned bilingual embeddings with bilingual lexicon induction (BLI).
The task is to retrieve the target translations with source words by searching for nearest neighbors with cosine similarity in the bilingual embedding space.
The evaluation metric used in prior work is usually top-k precision, but here we use a more informative measure, mean reciprocal rank (MRR) as recommended by \citet{Glavas:2019vv}.

\minisection{Fixed Target Context Window Settings.} First, we consider the settings where the target context size is fixed, and the source context size is configurable.
This setting assumes common situations where the embedding of the target language is available in the form of pre-trained embeddings.

Figure \ref{wiki_comp_fixed_window} shows the result of the four languages.
Firstly, we observe that too small windows (1 to 3) for source embeddings do not yield good performance, probably because the model failed to train accurate word embedding models with insufficient training word-context pairs that the small windows capture.

\begin{figure*}[t]
\begin{center}
  \includegraphics[width=15cm]{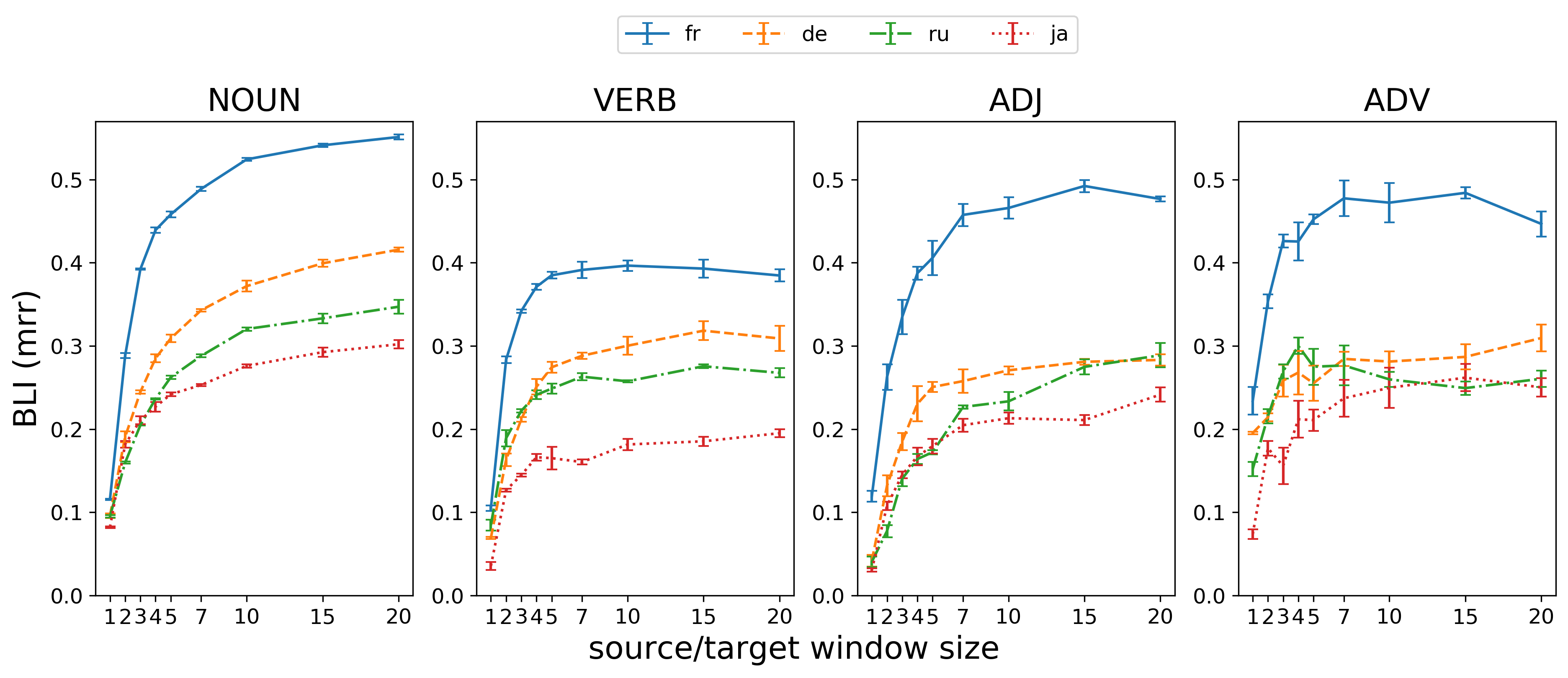}
  \caption{BLI performance for each PoS in the comparable setting.}
  \label{wiki_comp_pos}
\end{center}
\vspace{-5mm}
\end{figure*}

At first, this result may seem to contradict with the result from \citet{Sogaard:2018vo}. They trained English and Spanish embeddings with {\tt fasttext} \citep{Bojanowski:2017un} and the window size of 2, and then aligned them with an unsupervised mapping algorithm \citep{Lample:2018wg}. When they changed the window size of the Spanish embedding to 10, they only observed a very slight drop on top-1 precision (from 81.89 to 81.28). We suspect that the discrepancy with our result is due to the different settings. First of all, {\tt fasttext} adopts a dynamic window mechanism, which may obfuscate the difference in the context window. Also, they trained embeddings with full Wikipedia articles, which is an order of magnitude larger than ours; the {\tt fasttext} algorithm, which takes into account the character n-gram information of words, can exploit a non-trivial amount of subword overlap between the quite similar languages.

Overall, we observe that the best context window size for the source embeddings increases as the target context size increases, and increasing the context sizes of both the source and target embedding seems beneficial to the BLI performance.

\begin{figure}[t]
\begin{center}
  \includegraphics[width=7cm]{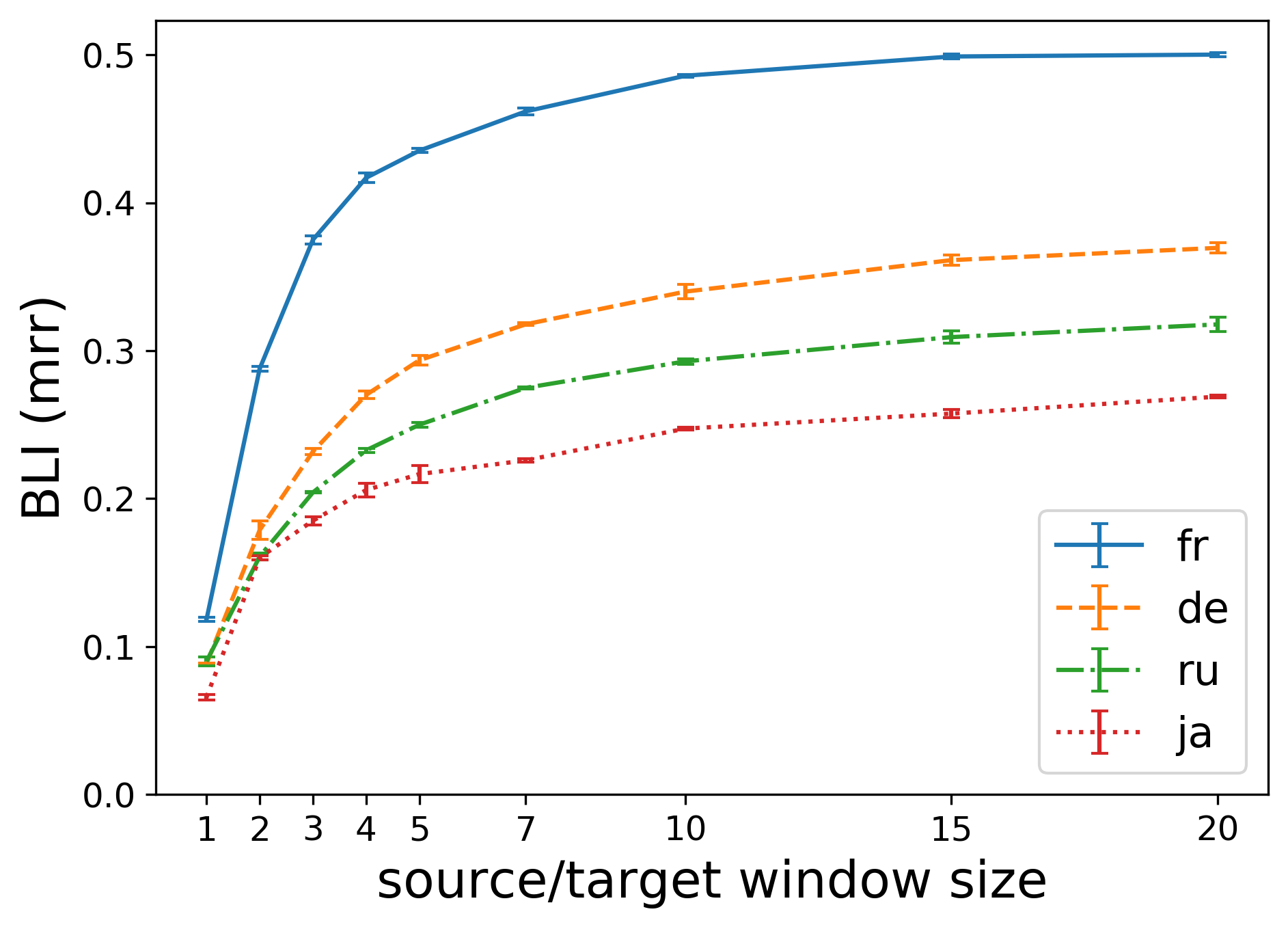}
  \caption{BLI performance in the comparable setting.
  }
  \label{wiki_comp_configurable}
\end{center}
\vspace{-5mm}
\end{figure}

\minisection{Configurable Source/Target Context Window Settings.} Hereafter, we present the results where both the source and target sizes are configurable and set to the same. Figure \ref{wiki_comp_configurable} summarizes the result of the same domain setting.

\begin{figure}[t]
\begin{center}
  \includegraphics[width=7cm]{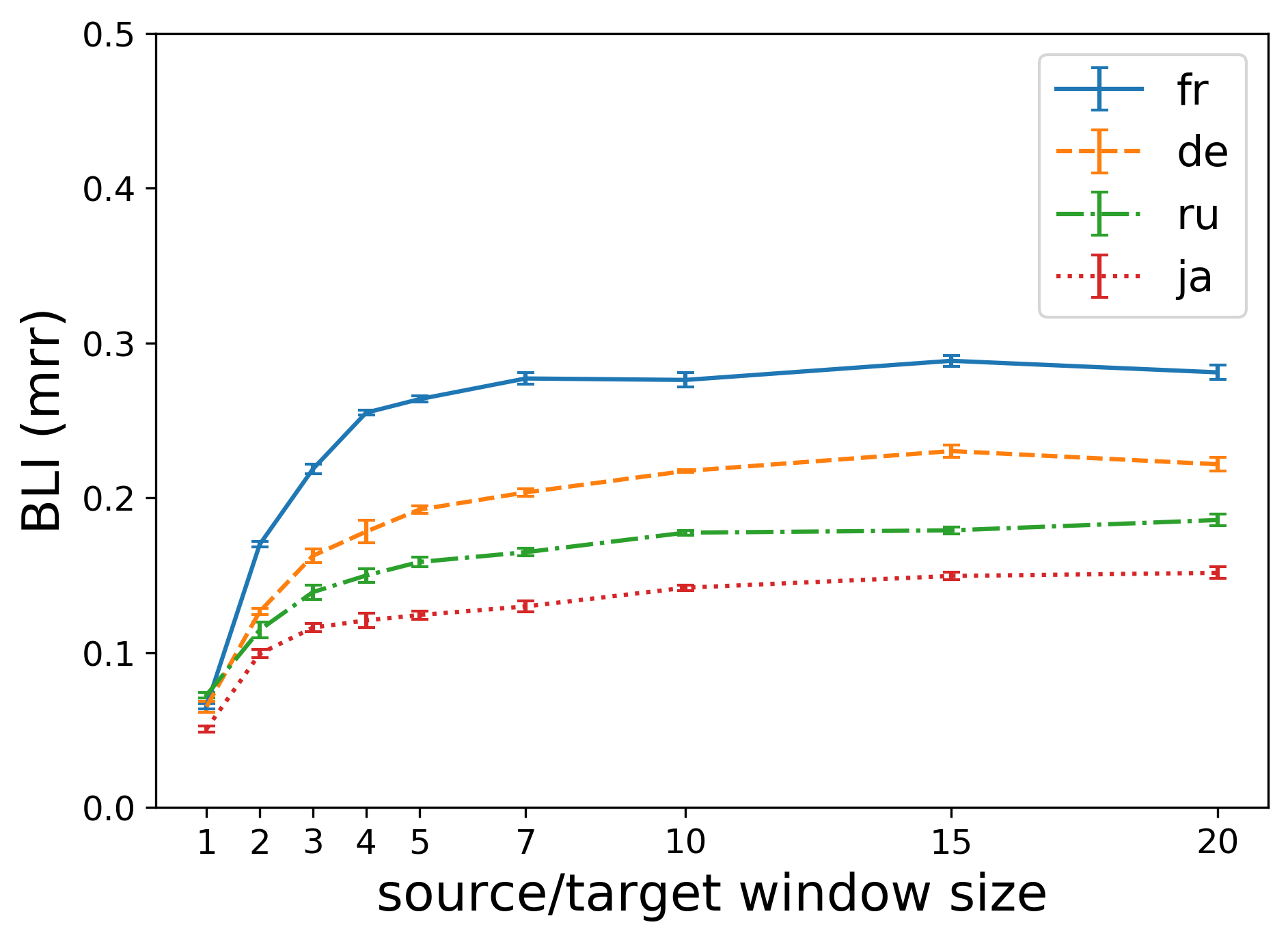}
  \caption{BLI performance in the different domain setting.
  }
  \label{news_configurable}
\end{center}
\vspace{-5mm}
\end{figure}

As we expected from the observation of the settings where the target window size is fixed, the performance consistently improves as the source and target context sizes increase.
Given that the larger context windows tend to capture topical similarities of words, we hypothesize that the more topical the embeddings are, the easier they are to be aligned.
Topics are invariant across different languages to some extent as long as the corpora are comparable. It is natural to think that topic-oriented embeddings capture language-agnostic semantics of words and thus are easier to be aligned among different languages.

This hypothesis can be further supported by looking at the metrics of each part-of-speech (PoS).
Intuitively, nouns tend to be more representative of topics than other PoS, and thus are expected to show a high correlation with the window size.
Figure \ref{wiki_comp_pos} shows the scores for each PoS. \footnote{We assigned to each word its most frequent PoS tag in the Brown Corpus \citep{kucera_computational_1967}, following \citet{wada-etal-2019-unsupervised}.}
In all languages, nouns and adjectives show stronger (almost perfect) correlation than verbs and adverbs.



\begin{figure*}[h]
\begin{center}
  \includegraphics[width=15cm]{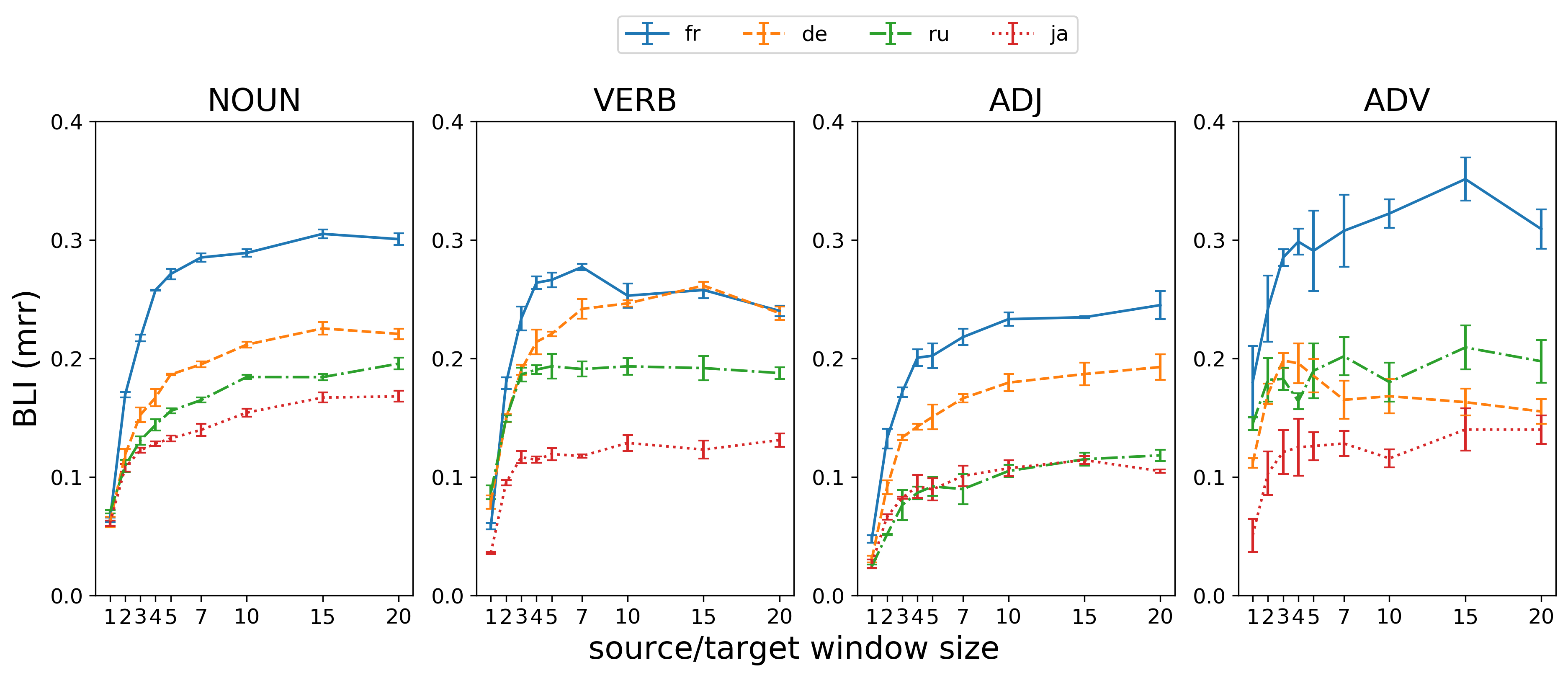}
  \caption{BLI performance for each PoS in the different domain setting.
  }
  \label{news_pos}
\end{center}
\vspace{-5mm}
\end{figure*}

\begin{figure}[t]
\begin{center}
  \includegraphics[width=7cm]{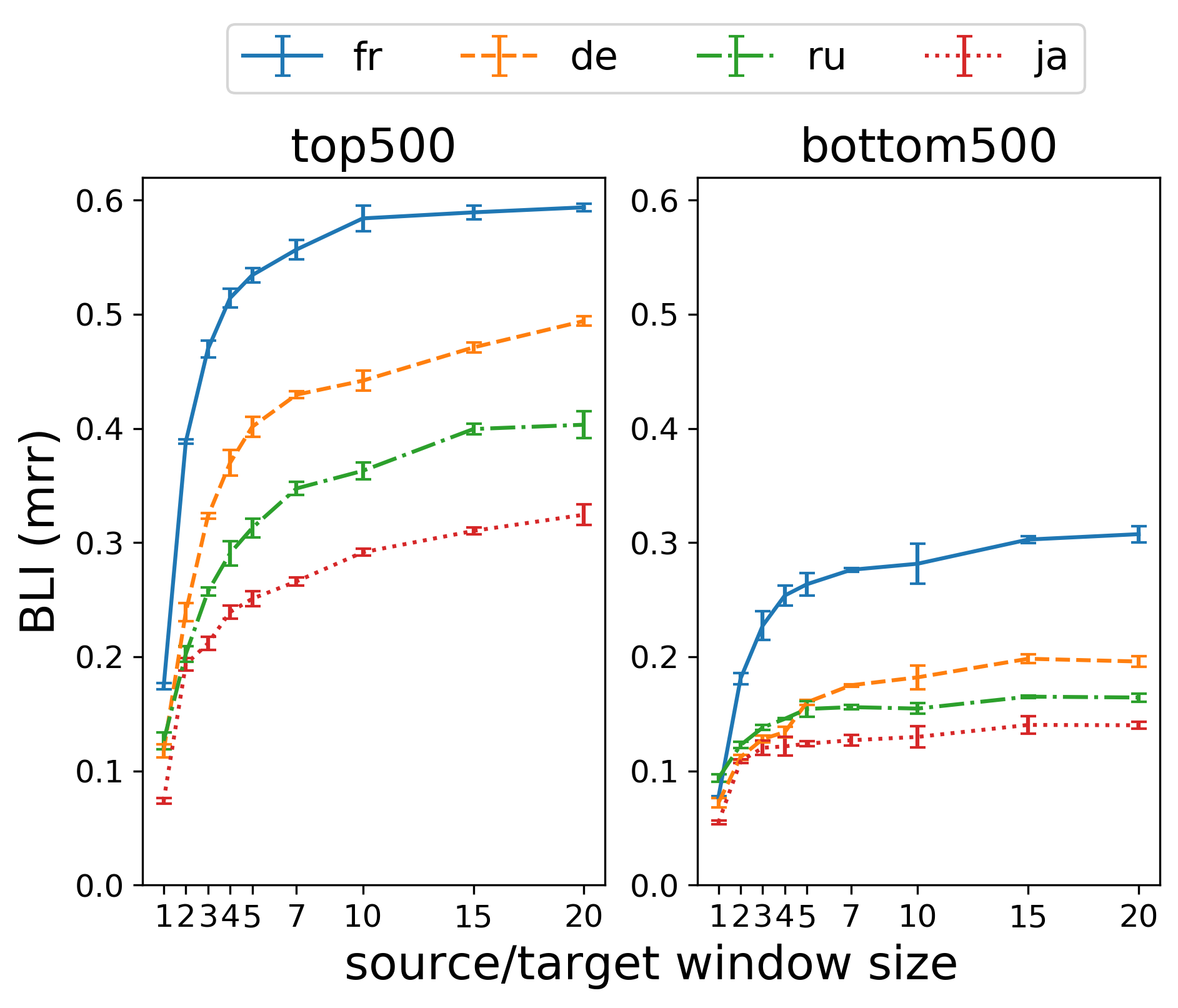}
  \caption{BLI performance with the top 500 frequent and rare words in the comparable setting.
  }
  \label{freq_analysis_same_domain}
\end{center}
\vspace{-5mm}
\end{figure}

\begin{figure}[t]
\begin{center}

  \includegraphics[width=7cm]{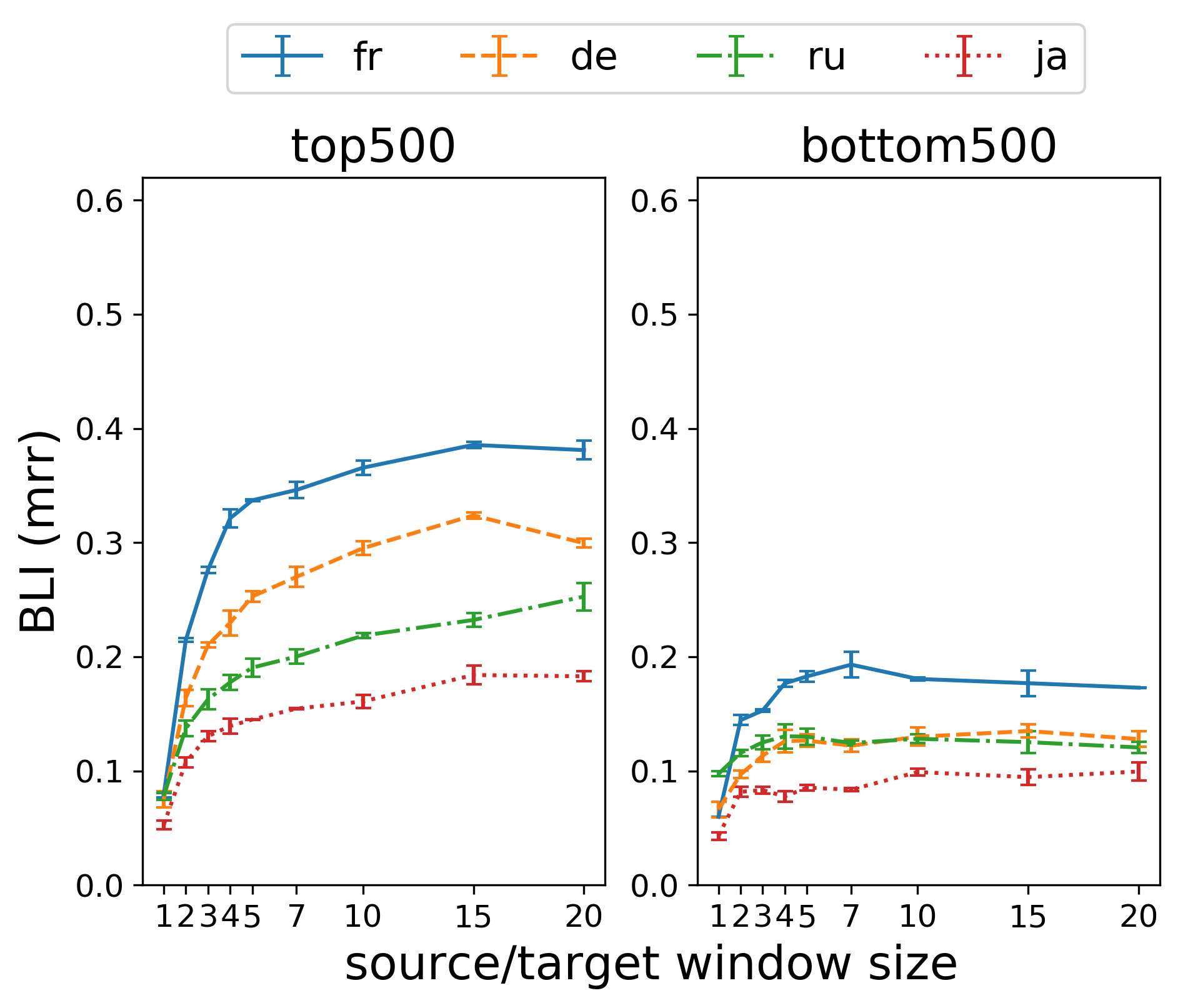}
  \caption{BLI performance on the top 500 frequent and rare words in the different domain setting.
  }
  \label{freq_analysis_different_domain}
\end{center}
\vspace{-5mm}
\end{figure}

\minisection{Different-domain Settings.} The results so far are obtained in the settings where the source and target corpora are comparable. When the corpora are comparable, it is natural that topical embeddings are easier to be aligned as comparable corpora share their topics. In order to see if the observations from the comparable settings hold true for different-domain settings, we also present the result from the different-domain (news) source corpora in Figure \ref{news_configurable}.

Firstly, compared to the same-domain settings (Figure \ref{wiki_comp_configurable}), the scores are lower by around 0.1 to 0.2 points across the languages and context windows, even with the same amount of training data.
This result confirms previous findings showing that domain consistency is important to the isomorphism assumption \citep{Sogaard:2018vo}.

As to the relation between the BLI performance and the context window, we observe a similar trend to the comparable settings: increasing the context window size basically improves the performance.
Figure \ref{news_pos} summarizes the results for each PoS. The performance on nouns and adjectives still accounts for much of the correlation with the window size.
This suggests that even when the source and target domains are different, some domain-invariant topics are captured by larger-context embeddings for nouns and adjectives.


\minisection{Frequency Analysis.} To further gain insight into what kind of words receive the benefit of larger context windows, we analyze the effect of word frequency. We extract the top and bottom 500 frequent words\footnote{The frequencies were calculated from our subset of the English Wikipedia corpus.} from the test vocabularies and evaluate the performance on them respectively.

The results of the comparable setting in each language are shown in Figure \ref{freq_analysis_same_domain}.

\begin{figure*}[t]
\begin{center}
  \includegraphics[width=15cm]{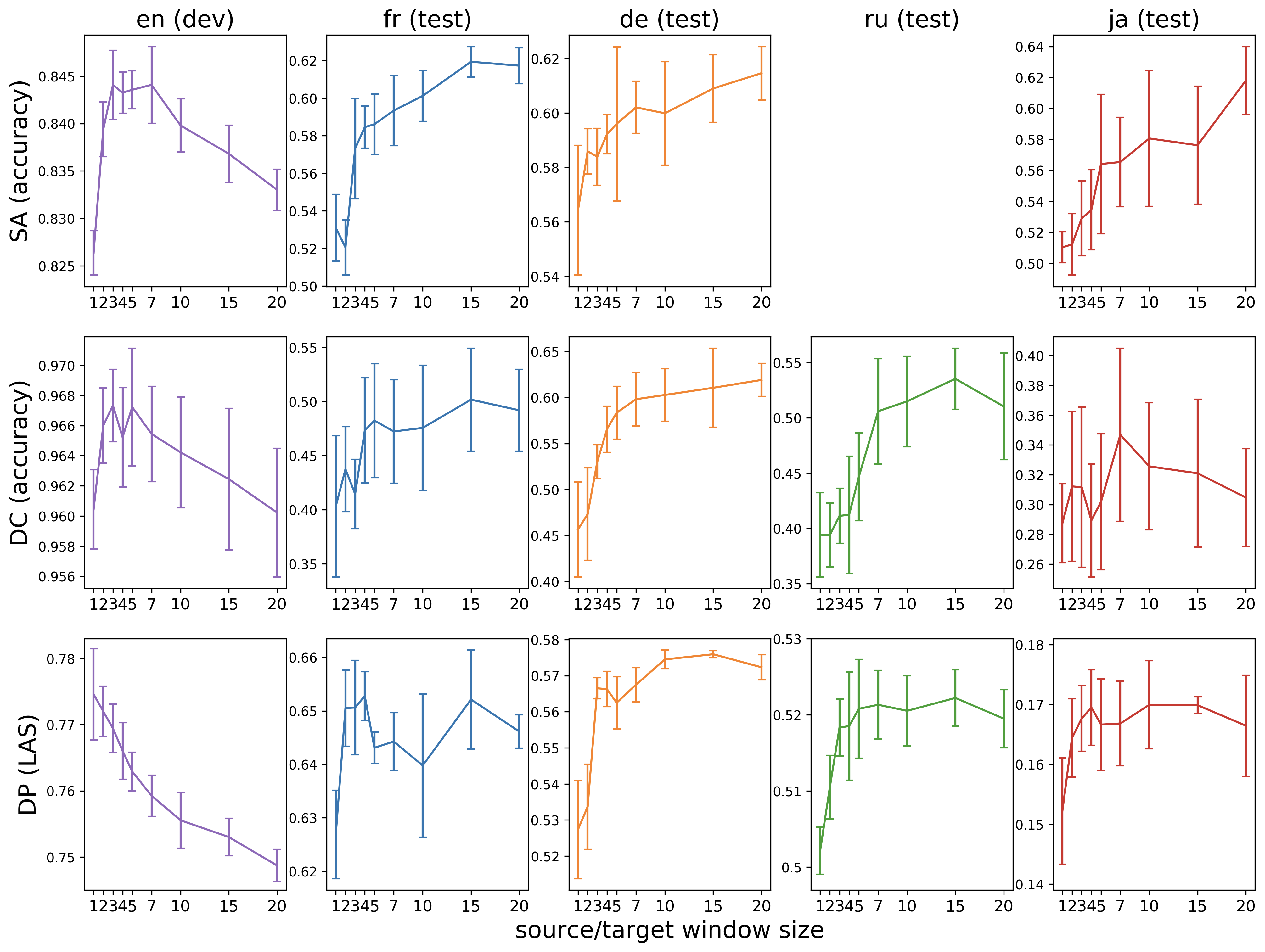}
  \caption{Downstream evaluations in the comparable settings. SA: sentiment analysis; DC: document classification; DP: dependency parsing. The window sizes of both the source and target embeddings are varied.}
  \label{downstream_results}
\end{center}
\vspace{-5mm}
\end{figure*}

The scores for the frequent words (top500) are notably higher than the rare words (bottom500). This confirms previous empirical results that existing mapping-based methods perform significantly worse for rare words \citep{Braune:2018wt, Czarnowska:2019vt}.

With respect to the relation with the context size, both frequent and rare words benefit from larger window sizes, although the gain in the rare words is less obvious in some languages ({\textsc ja} and {\textsc ru}).

In the different domain settings, as shown in Figure \ref{freq_analysis_different_domain}, the rare words, in turn, suffer from larger window sizes, especially for {\textsc fr} and {\textsc ru}, but the performance on frequent words still improves as the context window increases.

We conjecture that when training a skip-gram model, frequent words observe many context words, and that would mitigate the effect of irrelevant words (noise) caused by a larger window size and result in high-quality topical embeddings; however, rare words have to rely on a limited number of context words, and larger windows just amplify the noise and domain difference to result in an inaccurate alignment of them.

\section{Downstream Tasks}

Although BLI is a common evaluation method for bilingual embeddings, good performance on BLI does not necessarily generalize to downstream tasks \citep{Glavas:2019vv}.
To further gain insight into the effect of the context size on bilingual embeddings, we evaluate the embeddings with three downstream tasks: 1) sentiment analysis; 2) document classification; 3) dependency parsing. Here, we briefly describe the dataset and model used for each task.

\minisection{Sentiment Analysis (SA).} We use the Webis-CLS-10 corpus\footnote{\url{https://webis.de/data/webis-cls-10.html}} \cite{Prettenhofer:2010tt}, which is comprised of Amazon product reviews in the four languages: English, German, French, and Japanese (no Russian data available). We cast sentiment analysis as a binary classification task, where we label reviews with the scores of 1 or 2 as {\tt negative} and reviews with 4 or 5 as {\tt positive}.
For the model, we employ a simple CNN encoder followed by a multi-layer perceptrons classifier.

\minisection{Document Classification (DC).} MLDoc\footnote{\url{https://github.com/facebookresearch/MLDoc}} \cite{Schwenk:2018wqa} is compiled from the Reuters corpus for eight languages including all the languages used in this paper. The task is a four-way classification of the news article topics: {\tt Corporate/Industrial}, {\tt Economics}, {\tt Government/Social}, and {\tt Markets}. We use the same model architecture as sentiment analysis.

\minisection{Dependency Parsing (DP).} We train deep biaffine parsers \citep{Dozat:2017wp} with the UD English EWT dataset\footnote{\url{https://universaldependencies.org/treebanks/en_ewt/index.html}} \citep{Silveira:2014vs}. We use the PUD treebanks\footnote{\url{https://universaldependencies.org/conll17/}} as test data.

The hyperparameters used in this experiment are shown in Appendix \ref{appendix:DT}.

\minisection{Evaluation Setup.} We evaluate in a cross-lingual transfer setup how well the bilingual embeddings trained with different context windows transfer lexical knowledge across languages.
Here, we focus on the settings where both the source and target context sizes are varied.

For each task, we train models with our pre-trained English embeddings. We do not update the parameters of the embedding during training.
Then, we evaluate the model with the test data in other languages available in the dataset. At test time, we feed the model with the word embeddings of the test language aligned to the training English embeddings.

We train nine models in total for each setting with different random seeds and English embeddings, and we present their average scores and standard deviations.

\minisection{Result and Discussion.} The results from all the three tasks are presented in Figure \ref{downstream_results}.

For sentiment analysis and document classification, we observe a similar trend where the best window size is around 3 to 5 for the source English task, but for the test languages, larger context windows achieve better results.
The only deviation is the Japanese document classification, where the score does not show a significant correlation. We attribute this to low-quality alignments due to the large typological difference between English and Japanese, which can be confirmed by the fact that the Japanese scores are the lowest across the board.

For dependency parsing, embeddings with smaller context windows perform better in the source English task, which is consistent with the observation that smaller context windows tend to produce syntax-oriented embeddings \citep{Levy:2014we}. However, the performance of the small-window embeddings does not transfer to the test languages. The best context window for the English development data (the size of 1) performs the worst for all the test languages, and the transferred accuracy seems to benefit from larger context sizes, although it does not always correlate with the window size. This observation highlights the difficulty of transferring syntactic knowledge across languages. Word embeddings trained with small windows capture more grammatical aspects of words in each language, which, as different languages have different grammars, makes the source and target embedding spaces so different that it is difficult to align them.

In summary, a general trend we observe here is that good context windows in the source language task do not necessarily produce good transferrable bilingual embeddings. In practice, it seems better to choose a context window that aligns the source and target well, rather than using the window size that just performs the best for the source language.

\section{Conclusion and Future Work}
Despite their obvious connection, the relation between the choice of context window and the structural similarity of two embedding spaces has not been fully investigated in prior work.
In this study, we have offered the first thorough empirical results on the relation between the context window size and bilingual embeddings, and shed new light on the property of bilingual embeddings.
In summary, we have shown that:

\begin{itemize}
  \item larger context windows for both the source and target facilitate the alignment of words, especially nouns.
  \item for cross-lingual transfer, the best context window for the source task is often not the best for test languages. Especially for dependency parsing, the smallest context size produces the best result for the source task, but performs the worst for test languages.
\end{itemize}

We hope that our study will provide insights into ways to improve cross-lingual embeddings by not only mapping methods but also the properties of monolingual embedding spaces.

\section*{Acknowledgement}
We thank the anonymous reviewers for their valuable comments and suggestions. This work was supported by JST CREST Grant Number JPMJCR1513, Japan.

\newpage

\bibliography{references}
\bibliographystyle{acl_natbib}

\newpage
\onecolumn
\appendix
\section{The hyperparameters for training monolingual word embeddings}
\label{appendix:WE}

\begin{table}[h]
\begin{tabular}{lcc} \hline
hyperparameter             & \begin{tabular}[c]{@{}c@{}}Source embeddings\\ (1M sentences)\end{tabular} & \begin{tabular}[c]{@{}c@{}}Target embeddings\\ (5M sentences)\end{tabular} \\ \hline
embedding size             & \multicolumn{2}{c}{300}                                                                                                                                 \\
number of negative samples & \multicolumn{2}{c}{15}                                                                                                                                  \\
alpha (learning rate)      & \multicolumn{2}{c}{0.025 (linearly decayed during training)} \\
minimum word count         & 10                                                                         & 15                                                                         \\
number of iterations       & 10                                                                         & 5                                                                          \\ \hline

\end{tabular}
\end{table}

\section{The hyperparameters for downstream tasks}
\label{appendix:DT}

\subsection{Document Classification and Sentiment Analysis}
\begin{table}[h]
\begin{tabular}{clc} \hline
\multicolumn{2}{c}{hyperparameters}                    &            \\ \hline
\multirow{3}{*}{CNN Classifier} & number of filters    & 100        \\
                                & ngram\_filter\_sizes & 2, 3, 4, 5 \\
                                & MLP hidden size      & 64         \\ \hline
\multirow{4}{*}{Training}       & optimizer       & Adam                                                   \\
                                                  & learning rate          & 0.001 (halved each time the dev score stops improving) \\
                                                  & patience               & 3                                                      \\
                                                  & batch size             & 64               \\ \hline
\end{tabular}
\end{table}

\subsection{Dependency Parsing}
\begin{table}[h]
\begin{tabular}{clc} \hline
\multicolumn{2}{c}{hyperparameters}                           & \multicolumn{1}{r}{}                                   \\ \hline
\multirow{5}{*}{Graph-based Parser}   & LSTM hidden size       & 200                                                    \\
                                     & LSTM number of  layers & 3                                                      \\
                                     & tag representation dim & 100                                                    \\
                                     & arc representation dim & 500                                                    \\
                                     & pos tag embedding dim  & 50                                                     \\ \hline
\multirow{4}{*}{Training}            & optimizer              & Adam                                                   \\
                                     & learning rate          & 0.001 (halved each time the dev score stops improving) \\
                                     & patience               & 3                                                      \\
                                     & batch size             & 32 \\ \hline
\end{tabular}
\end{table}

\end{document}